\documentclass[letterpaper, 10 pt, conference]{ieeeconf}  

\IEEEoverridecommandlockouts                              
\usepackage[T1]{fontenc}
\usepackage[utf8]{inputenc}
\usepackage{authblk}
\usepackage{graphicx}
\usepackage{subfigure}
\usepackage{array}
\usepackage{float}
\usepackage{float}
\usepackage{booktabs}
\usepackage{bbding}
\usepackage{etoolbox}
\usepackage{color}
\usepackage{soul}
\usepackage{algorithm}
\usepackage{algorithmic}
\usepackage{cite}
\usepackage{balance}
\usepackage{amsmath, amssymb}
\usepackage{makecell}
\usepackage{multirow}
\usepackage{hyperref}

\title{\LARGE \bf
Deep Reinforcement Learning-Based Control for Stomach Coverage Scanning of Wireless Capsule Endoscopy
}

\author{Yameng Zhang$^{*1}$, Long Bai$^{*1}$, 
Li Liu$^{1}$, Hongliang Ren$^{2}$, Max Q.-H. Meng$^3$, \emph{Fellow, IEEE}
\thanks{*This work was supported by Hong Kong Research Grants Council (RGC) Collaborative Research Fund (CRF C4026-21GF and CRF C4063-18G), and General Research Fund (GRF \#14211420 and GRF \#14204321); the National Key R\&D Program of China under Grant 2018YFB1307700 (also with subprogram 2018YFB1307703) from the Ministry of Science and Technology (MOST) of China; Shun Hing Institute of Advanced Engineering (BME-p1-21/8115064) at the Chinese University of Hong Kong (CUHK). \{\emph{Corresponding to: Li Liu, Hongliang Ren.}\}}
\thanks{$*$Equal Contribution.}
\thanks{$^{1}$Y. Zhang, L. Bai, and L. Liu are with the Department of Electronic Engineering, The Chinese University of Hong Kong, Hong Kong. (email: zhangyameng@link.cuhk.edu.hk, b.long@ieee.org, liliu@cuhk.edu.hk)}%
\thanks{$^{2}$Hongliang Ren is with the Department of Electronic Engineering, The Chinese University of Hong Kong, Hong Kong, and the Shun Hing Institute of Advanced Engineering, The Chinese University of Hong Kong, Hong Kong; also with the Department of Biomedical Engineering, National University of Singapore, Singapore, and NUS (Suzhou) Research Institute, Suzhou 215000, China. (email: hlren@ieee.org)} 
\thanks{$^{3}$Max Q.-H. Meng is with the Shenzhen Key Laboratory of Robotics Perception and Intelligence, and the Department of Electronic and Electrical Engineering, Southern University of Science and Technology, Shenzhen 518055, China; on leave from the Department of Electronic Engineering, The Chinese University of Hong Kong, Hong Kong, and from the Shenzhen
Research Institute, The Chinese University of Hong Kong at Shenzhen, Shenzhen 518057, China. (email: max.meng@ieee.org)}
}

\begin{document}
\maketitle
\thispagestyle{empty}
\pagestyle{empty}

\begin{abstract}

Due to its non-invasive and painless characteristics, wireless capsule endoscopy has become the new gold standard for assessing gastrointestinal disorders. Omissions, however, could occur throughout the examination since controlling capsule endoscope can be challenging. In this work, we control the magnetic capsule endoscope for the coverage scanning task in the stomach based on reinforcement learning so that the capsule can comprehensively scan every corner of the stomach. We apply a well-made virtual platform named VR-Caps to simulate the process of stomach coverage scanning with a capsule endoscope model. We utilize and compare two deep reinforcement learning algorithms, the Proximal Policy Optimization (PPO) and Soft Actor-Critic (SAC) algorithms, to train the permanent magnetic agent, which actuates the capsule endoscope directly via magnetic fields and then optimizes the scanning efficiency of stomach coverage. We analyze the pros and cons of the two algorithms with different hyperparameters and achieve a coverage rate of 98.04\% of the stomach area within 150.37 seconds.

\end{abstract}

\section{INTRODUCTION}
\begin{figure}[t] 
    \centering
    \includegraphics[width=7.5cm]{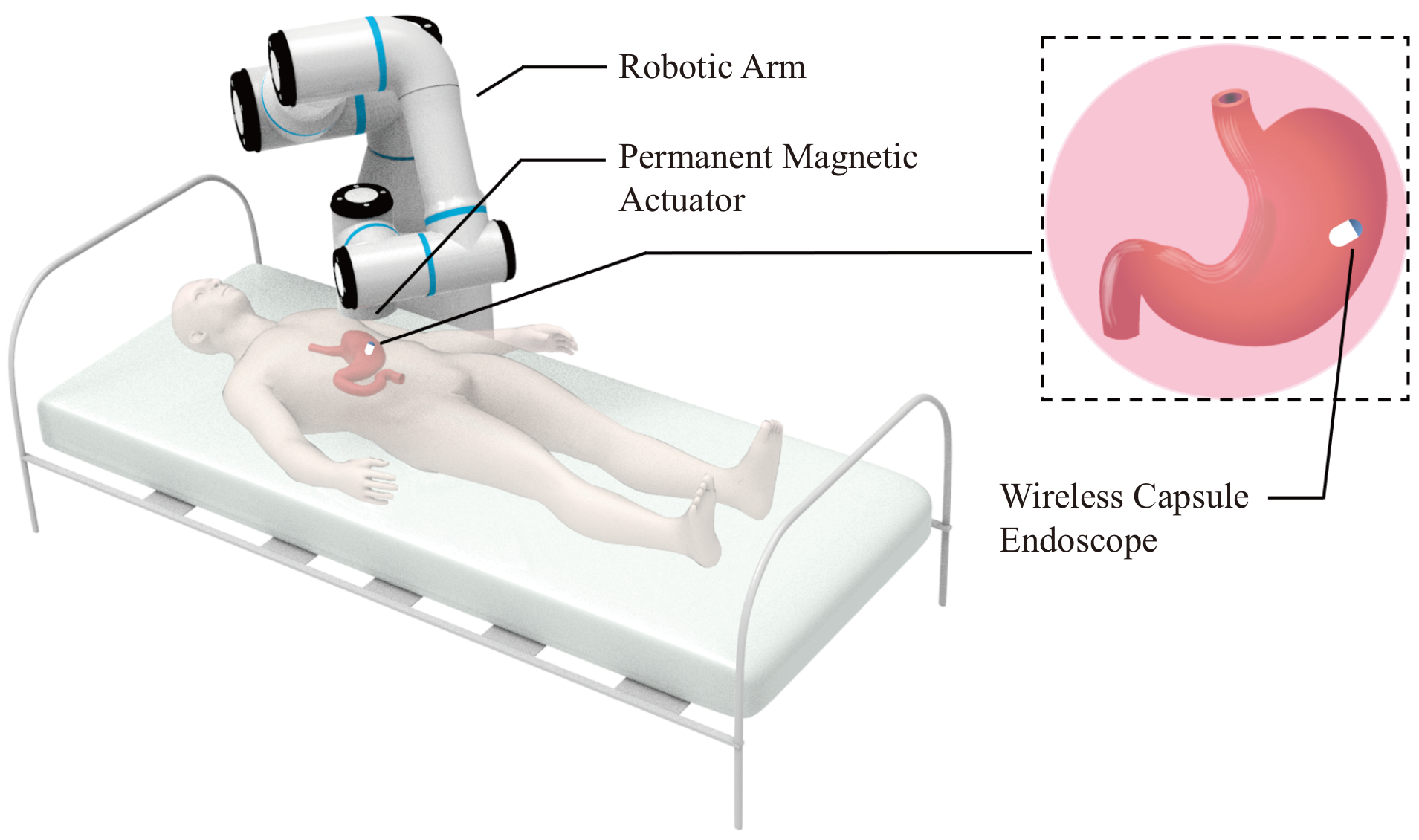}
    \caption{Overview of magnetically controlled WCE system.
    }
    \label{fig:1}
\end{figure}

As shown in Figure \ref{fig:1}, wireless capsule endoscope (WCE) is a swallowable intelligent capsule robot for \textit{in vivo} imaging, examination, and diagnosis of the gastrointestinal tract. Compared with normally used gastroscopy and colonoscopy, WCE excels for its non-invasive and less painful examination and diagnosis~\cite{c1,c6}. WCE can also implement many other tasks such as PH measurement, temperature detection, biopsy, precision drug delivery, etc.~\cite{c2,c15}, which offer physicians more comprehensive and valuable information. However, all those assignments rely on dynamic adjustment and control of WCE. The remote operation of WCE is very challenging since its localizing and controlling are full of uncertainty and complexity~\cite{c5,c10}. Many researchers are endeavoring to develop artificial intelligence models for the motion control of WCE.

Currently, machine learning (ML) has achieved extensive success in medical field~\cite{c20,c21,c23,c24,c25}. Reinforcement learning (RL), one of the paradigms and methodologies of ML, is used to describe and solve the problem of maximizing the reward or achieving a specific goal by learning strategies during the agent's interaction with the environment. It has been effectively used in many domains, including computer games~\cite{c12}, robotic arm control~\cite{c13}, autonomous driving~\cite{c9}, etc. The agent must rely on its own experiences to learn from and enhance its course of action to adapt to the environment because the external world offers little information~\cite{c8}. 
However, real-world systems always have high-dimensional input data. It might be difficult for RL algorithms to select the proper action. To effectively handle the high-dimensional input data, we often utilize the neural network to map the high-dimensional observation data to our desired action data. Thus, when a neural network is applied in the RL algorithm, we call it the deep reinforcement learning (DRL) algorithm~\cite{c4}. 

In this work, we use DRL algorithms to train a permanent magnetic agent, which actuates WCE directly via the magnetic field and then optimizes the efficiency of stomach coverage scanning. The critical points of our work are listed below:

\begin{itemize}

\item A permanent magnet will be trained and used to control the pose and motion of WCE through the magnetic field. Then, the WCE with a monocular camera will automatically scan the maximum coverage area of the stomach within minimal operating time.

\item We compare the effectiveness of two DRL algorithms, i.e., the PPO~\cite{c14} and SAC~\cite{c3}, in training the permanent magnetic agent. We analyze the effect of the learning rate on the training results; hence we get a set of appropriate hyperparameters for the training process. 

\end{itemize}

\section{RELATED WORK}
\label{sec2}
Researchers have used RL algorithms to train WCE in virtual, artificial, or real environments. G. Trovato \textit{et al.}~\cite{c19} developed a robotic endoscope with flexible helical fins and controlled its locomotion through Q-learning and SARSA algorithms. The device was proven capable of passing through colons at an appropriate speed without getting stuck. L. Wu \textit{et al.}~\cite{c18} constructed a real-time quality improving system, WISENSE, to enhance endoscopy's scanning quality and efficiency. The system was developed with deep convolutional neural networks (DCNN) along with DRL algorithms and compared with a control group. Experiments proved that WISENSE significantly reduced the blind spot rate of the esophagogastroduodenoscopy procedure. M. Turan \textit{et al.}~\cite{c16} used a DRL algorithm to learn the continuous control of a magnetically actuated soft capsule endoscope. Their controller could reduce the tedious physical modeling of complex and highly nonlinear behaviors. A. Marino \textit{et al.}~\cite{c11} applied an RL-based control method to the magnetically flexible endoscope to reduce pain and increase ergonomics in colonoscopy. They kept the endoscope in contact with the tissue while ensuring that the endoscope automatically navigated the entire colon. K. İncetan \textit{et al.}~\cite{c7} created a virtual gastrointestinal environment, VR-Caps, to study the possibilities of applying various machine learning algorithms, providing an easy-to-use simulation platform for us to develop approaches that improve the quality of the gastrointestinal treatment.

\section{METHODOLOGY}
\label{sec3}
We use the ML-Agents toolkit~\cite{c17} to implement our training objective. ML-Agents toolkit is an open-source package built in the Unity environment. It provides some implementations of state-of-the-art DRL algorithms, which is pretty helpful in training intelligent agents for virtual games. However, before we start training, we have to define and formulate some primary constraints and policies in the environment so that the agent is trained within our expectations. Here are two critical problems: how we formulate the scanning level in the stomach and how we define the reward in the training episode, so in Sections~\ref{sec3-1} and~\ref{sec3-2}, we describe our solutions to the two problems, respectively. In Section~\ref{sec3-3}, we give an introduction to the PPO and SAC algorithms. 

\subsection{Scanning Level Definition}
\label{sec3-1}
In the environment setup, we assign a monocular camera to the WCE so that the WCE can obtain the inner images of the stomach in real time. However, this does not work for the measurement of the scanning area. To solve this problem, we use the point cloud chart of the stomach to formulate the scanning level. The underlying logic is that we fix the perspective angle of the WCE camera. When the camera captures vertices, we accumulate the number of vertices having been scanned and calculate the real-time coverage ratio by dividing the number of visible vertices by the number of all vertices. This method helps measure the coverage ratio difference between two consecutive steps and is thus pretty valuable in defining the reward list. We provide the pseudo-codes of the vertice detection in Algorithm \ref{alg1}, where more detailed processes can be found.

\begin{algorithm} [t]
	\caption{Vertice detection in each episode} 
	\label{alg1}
	\begin{algorithmic}[1]
	    \REQUIRE point cloud chart of organ models
        \ENSURE $diff\_coverage$ in every step
		\STATE \textbf{Intialization:}
		\STATE $vertice\_count \gets 24822;$
		\STATE $visible\_vertices[~ ] \gets [ ~];$
		\STATE $color\_vertices[ ~] \gets [red];$
		\STATE $vertices[ ~] \gets vertice\_positions;$
		\STATE $visible\_count \gets 0$,
		$previous\_coverage\gets 0$,
		$step \gets 0,$
		$current\_coverage\gets 0$,
		$diff\_coverage \gets 0$;
		\WHILE{$step < 1500$}
		\STATE $step \gets step+1;$
		\STATE \textbf{Step with PPO or SAC algorithm};
		\STATE $i \gets 0;$
		\REPEAT
		\STATE $i \gets i+1;$
		\IF{$vertices[i]$ is visible for the camera}
		\STATE $visible\_vertices.append(vertices[i]);$
		\STATE $visible\_count \gets visible\_count +1;$
		\STATE $color\_vertices[i] \gets blue;$
		\ENDIF
		\UNTIL $i \geq vertice\_count$
		\STATE $current\_coverage \gets 100 \times \frac{visible\_count }{vertice\_count};$
		\STATE $diff\_coverage \gets current\_coverage-previous\_coverage;$
		\STATE $previous\_coverage \gets current\_coverage;$
		\ENDWHILE 
	\end{algorithmic} 
\end{algorithm}

\subsection{Reward Definition}
\label{sec3-2}

\begin{figure}[t] 
    \centering
    \includegraphics[width=7cm]{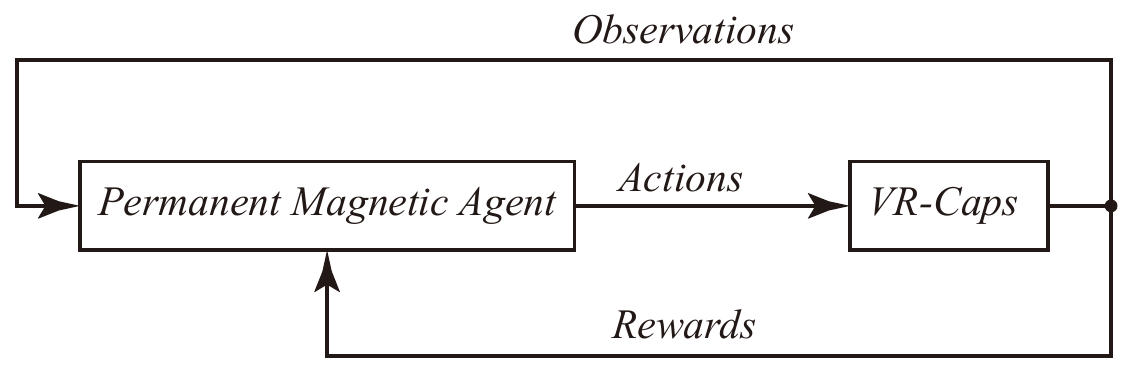}
    \caption{Training framework of DRL-based WCE system.
    }
    \label{fig:2}
\end{figure}

\begin{figure}[t]  
    \centering
    \includegraphics[width=8.5cm]{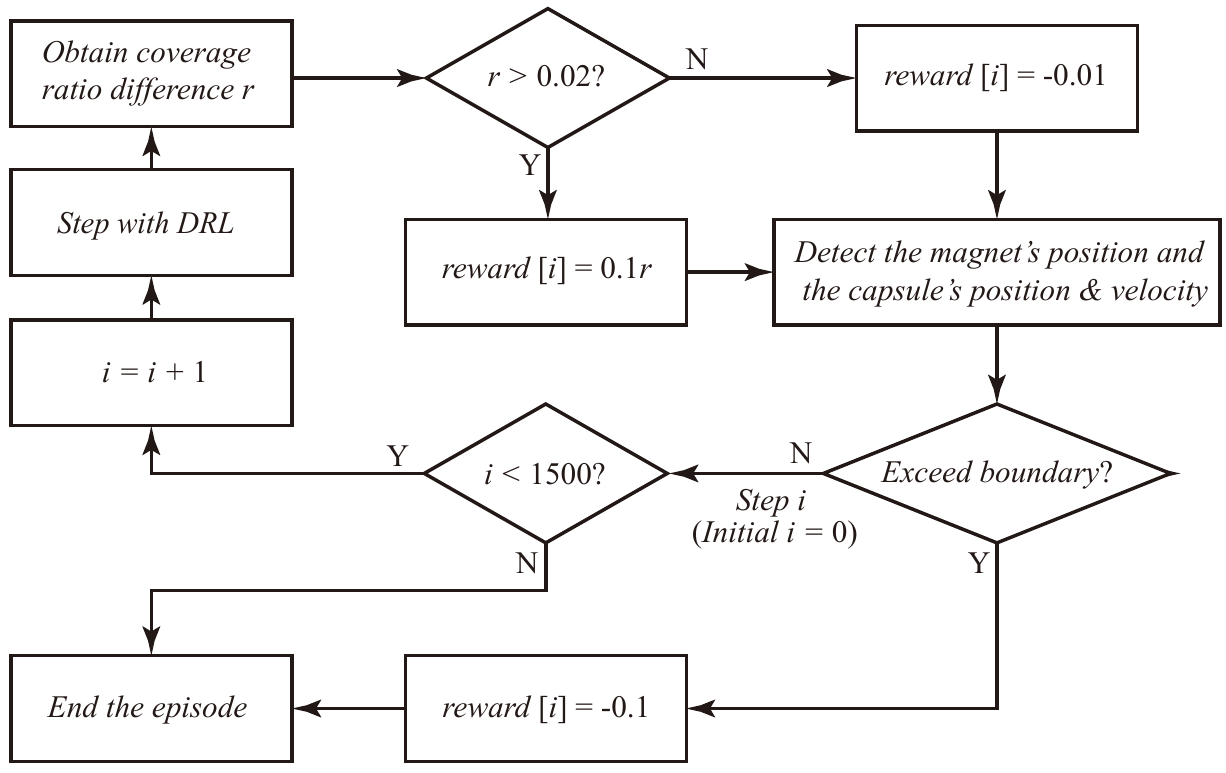}
    \caption{Training workflow of DRL-based WCE system within one episode.
    }
    \label{fig:3}
\end{figure}

The training framework of the DRL-based WCE system is presented in Figure \ref{fig:2}. The permanent magnetic agent takes actions in the VR-Caps environment, which provides an interpretation of reward and observations to the DRL algorithm, which then offers guidance on the agent's actions. The PPO and SAC algorithms are explicit DRL algorithms that are relatively easy to follow. But the environment is always so intricate and versatile that unexpected problems may happen with the failure of parameters formulation. In this project, we define the WCE's position, rotation, velocity, and the magnet's position as observations. The actions are defined as the two horizontally translational motions of the magnet. The reward for the coverage policy is the area coverage difference between two consecutive time steps, that is:
\begin{equation}
r_{t}=k (C_{t}-C_{t-1}),
\end{equation}
where $C_t$ denotes the area coverage ratio at time $t$. $k$ denotes a scaling factor for fast convergence. The area coverage ratio is calculated as the proportion of the vertices detected by the WCE's mono camera to all the vertices contained in the stomach. Besides, we assign penalty rewards for the agent when the WCE scans areas that have been detected. Also, when the capsule or magnet exceeds their position or velocity boundary, we should give a minus reward to the agent and then end the episode. We list the reward-defining pseudo-codes in Algorithm \ref{alg2}, which shows more details in reward calculation. Combining Algorithms \ref{alg1} and \ref{alg2}, the overall training workflow of the DRL-based WCE system is illustrated in Figure \ref{fig:3}, which shows detailed training processes within one episode with $1500$ steps at most.

\begin{algorithm}[t] 
	\caption{Magnetic agent training in each episode} 
	\label{alg2}
	\begin{algorithmic}[1]
		\REQUIRE position and velocity of the capsule and magnet
        \ENSURE reward in every step
		\STATE \textbf{Intialization:}
		\STATE $capsule\_angular\_velocity\gets 0, capsule\_velocity \gets 0;$
		\STATE $magnet\_angular\_velocity\gets 0,magnet\_velocity \gets 0;$
		\STATE $reward[~]\gets 0,step\gets 0,diff\_coverage\gets 0, r \gets 0; $
        \WHILE{$step < 1500$}
		\STATE \textbf{Collect Observations:}
		\STATE $capsule\_position,capsule\_rotation,capsule\_velocity$,\\$magnet\_position, magnet\_rotation \gets Unity;$
		\STATE $step \gets step+1;$
		\STATE \textbf{Step with PPO or SAC algorithm};
		\STATE $r \gets diff\_coverage;$
		\IF{$r>0.02$}
		\STATE $reward[step] \gets 0.1 \times r;$
		\ELSE
		\STATE $reward[step] \gets - 0.01;$
		\ENDIF
		\STATE \textbf{Action delivery:}
		\STATE $[magnet\_velocity, ~magnet\_angular\_velocity] \gets$ trained neural network;
		\STATE $magnet\_position \gets magnet\_position + magnet\_velocity \times \Delta t $;
		\STATE $magnet\_rotation \gets magnet\_rotation + magnet\_angular\_velocity \times \Delta t $;
		\IF{$capsule\_velocity$ exceeds its boundary}
		\STATE $reward[step] \gets - 0.1;$
		\STATE end the episode;
		\ELSIF{$capsule\_position$ exceeds its boundary}
		\STATE $reward[step] \gets - 0.1;$
		\STATE end the episode;
		\ELSIF{$magnet\_position$ exceeds its boundary}
		\STATE $reward[step] \gets - 0.1;$
		\STATE end the episode;
		\ENDIF
		
		\ENDWHILE 
	\end{algorithmic} 
\end{algorithm}

\subsection{DRL Method Description}
\label{sec3-3}
We select two widely used DRL algorithms, i.e., the PPO and SAC, to train the magnetic agent. Both the PPO and SAC algorithms are stochastic policy-based approaches. The PPO algorithm is an on-policy method. It has the stability and reliability of trust-region methods and is simple to implement. The SAC algorithm is an off-policy method. It has high sample efficiency and incorporates the clipped double-Q trick. Because of the inherent stochasticity of the policy in SAC, it benefits from something like target policy smoothing. 

During the training process, the agent is trained with many episodes, and the reward of every step is recorded until the cumulative reward converges to a point. Table \ref{table1} shows some hyperparameters we use in the DRL project. By adjusting the hyperparameters, such as the learning rate, we will see the effect of some hyperparameters on the training. Also, by comparing the reward and loss curves of the PPO and SAC, we can find and select an optimal training model for the capsule endoscope platform. Finally, we expect the area coverage ratio to increase dramatically and the detecting time to be shortened significantly.
\begin{table}[htbp]
\caption{Hyperparameters in the PPO and SAC algorithms}
\centering
\begin{tabular}{@{}lll@{}}
\toprule
Trainer & PPO & SAC \\ \midrule
Batch Size & 512 & 512 \\
Buffer Size & 4096 & 512000 \\
Hidden Units & 128 & 128 \\
Learning Rate & 0.001 & 0.0005 \\
Learning Rate Schedule & Linear & Linear \\
Max Steps & 3000000 & 3000000 \\
Memory Size & 128 & 128 \\
Num Layers & 2 & 2 \\
Time Horizon & 1024 & 1024 \\
Sequence Length & 64 & 64 \\
Summary Freq & 10000 & 10000 \\
Gamma & 0.99 & 0.99 \\
Num Epoch & 5 & N/A \\
Lambd & 0.95 & N/A \\
Epsilon & 0.2 & N/A \\
Beta & 0.005 & N/A \\
Tau & N/A & 0.005 \\ \bottomrule
\label{table1}
\end{tabular}
\end{table}

\section{SIMULATION VALIDATION}
\label{sec4}

\subsection{Environment Setup}

\begin{figure}[thbp]  
    \centering
    \includegraphics[width=8.2cm]{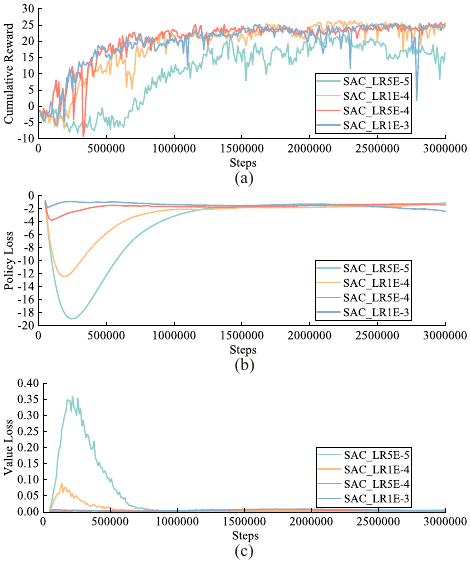}
    \caption{Simulation results of SAC. (a) cumulative reward; (b) policy loss; (c) value loss.
    }
    \label{fig:4}
\end{figure}

\begin{figure}[bhtp]   
    \centering
    \includegraphics[width=8.2cm]{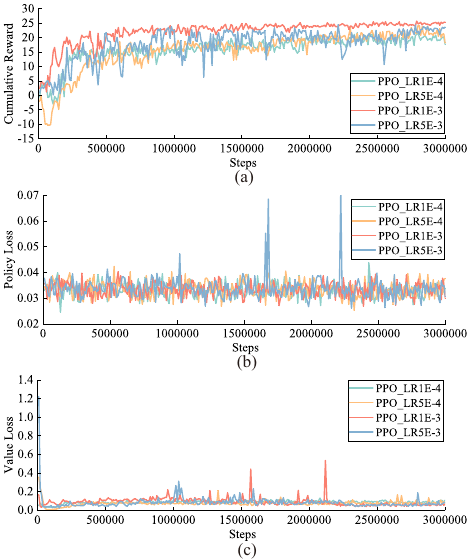}
    \caption{Simulation results of PPO. (a) cumulative reward; (b) policy loss; (c) value loss.
    }
    \label{fig:5}
\end{figure}

\begin{figure}[bhtp]   
    \centering
    \includegraphics[width=8.2cm]{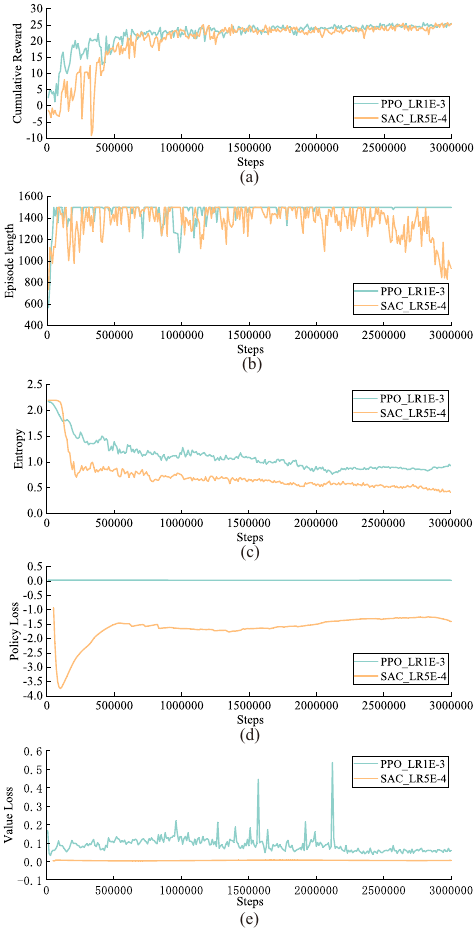}
    \caption{Comparison between PPO and SAC. (a) cumulative reward; (b) episode length; (c) entropy; (d) policy loss; (e) value loss.
    }
    \label{fig:6}
\end{figure}

We want to train a permanent magnetic agent to control the pose and motion of WCE via the magnetic field so that the WCE scans the maximum coverage area of the stomach within minimal operating time. However, it is hard for us to access patients and WCE devices in the real world. We utilize the Unity-based comprehensive simulation platform for WCE operations named VR-Caps \cite{c7} to facilitate our research. VR-Caps is a virtual active capsule environment that can simulate a series of realistic tissues, organs, and different types of WCE. Its virtual environment can help accelerate the design, testing, and optimization process of WCE robots. Besides, it is relatively easy to adjust the attributes and parameters of the VR-Caps environment. We use a mono camera WCE in VR-Caps and test it in a virtual stomach with viscosity and gravity. Moreover, we add a permanent magnet outside the stomach that can support the active movement of WCE. The virtual platform of our simulation is Unity Version 2019.3.2f1 and ML-Agents Release 1. The training environment is based on the Intel Core™ i7-12700KF CPU and NVIDIA RTX 3060Ti GPU. 

\subsection{Simulation Results}

The learning rate plays a significant role in both the model's final results and convergence speed. With a minimal learning rate, the loss of the model falls and converges very slowly. With a considerable learning rate, the parameters update so dramatically that the model will converge to a local optimum point, or the loss will increase instead. Therefore, in the respective simulation of SAC and PPO, we use four different learning rates preselected within the convergence range for training and seeking the best performance. The other parameters for training can be found in Table~\ref{table1}. The simulation results of SAC are shown in Figure~\ref{fig:4}. From the reward curves in Figure~\ref{fig:4} (a), we can see that the model converges the most slowly and the final reward (around 21) is the lowest when the learning rate is $5e^{-5}$, while the reward converges faster and achieves a higher value (around 26) at the learning rates of $1e^{-3}$, $5e^{-4}$, and $1e^{-4}$. On the other hand, the reward curves trained with the learning rates of $1e^{-3}$ and $1e^{-4}$ show unstable fluctuations. Therefore, the results of the $5e^{-4}$ learning rate can be considered the best results of the SAC algorithm. Meanwhile, The convergence speed of policy loss and value loss in Figure \ref{fig:4} becomes slower with decreasing learning rates, while the final losses are similarly low at different learning rates.

\begin{figure}[hbtp] 
    \centering
    \includegraphics[width=5cm]{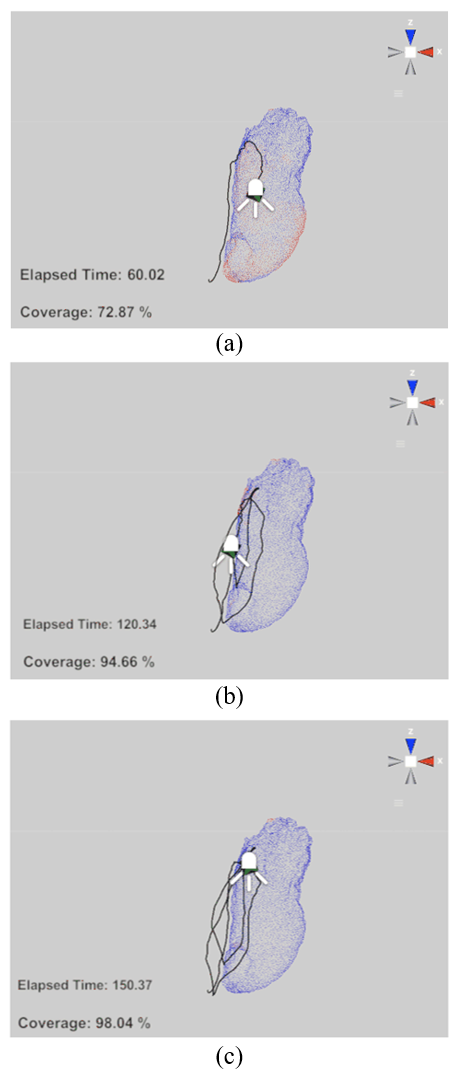}
    \caption{Coverage results of DRL in (a) 60 seconds, (b) 120 seconds, and (c) 150 seconds. The final coverage rate is 72.87\%, 94.66\%, and 98.04\% of the stomach area within 60.02, 120.34, and 150.37 seconds, respectively.
    }
    \label{fig:7}
\end{figure} 

Figure \ref{fig:5} shows the PPO algorithm trained with four different learning rates. Through the convergence results of the reward curves, the highest final reward value is the learning rate of $1e^{-3}$, followed by the learning rates of $5e^{-3}$, $5e^{-4}$, and $1e^{-4}$. Besides, the learning rate $1e^{-3}$ also achieves the best convergence speed. In policy loss and value loss, the learning rates of $1e^{-3}$, $5e^{-4}$, and $1e^{-4}$ do not differ much, and the $5e^{-3}$ learning rate shows significant instability in both the reward and loss curves. Therefore, the learning rate of $1e^{-3}$ should be the best result of the PPO algorithm.

Figure \ref{fig:6} shows the comparison between the best results of the PPO and SAC. Although the PPO presents a faster convergence rate, the final results on reward curves are similar between the PPO and SAC. The policy loss and value loss of the SAC are lower than those of the PPO. The difference in the entropy curves should come from the $tanh$ in the SAC policy, which ensures that policy output is bounded to a finite range. In this case, the SAC policy distribution is no longer the original Gaussian distribution. Conversely, this $tanh$ is absent in the PPO policy. In other words, the PPO calculates the entropy of old and new strategies using the actions before the $tanh$. In contrast, the SAC calculates the entropy of strategies using the actions after the $tanh$, leading to the difference in the entropy curves. Meanwhile, the SAC adds the derivative term of $tanh(a)$ when calculating the strategy entropy to offset the influence of $tanh$ as a correction.

\subsection{Comparison between Manual Control and DRL Control}

The DRL-based control is designed to give WCE a more comprehensive scanning efficiency while freeing human labor. In this case, we would like to compare the results between the manual control and our DRL-based control method. We trained 6 million steps with the SAC algorithm with a learning rate of $5e^{-4}$. Figure \ref{fig:7} shows the results of its trained model and its coverage rate at roughly 60 seconds, 120 seconds, and 150 seconds. Meanwhile, we invited an experienced WCE researcher to operate the capsule endoscope manually and recorded his coverage rate simultaneously. Table \ref{table2} shows the coverage comparison between the manual and DRL control. We know that the DRL-based WCE scanning is much more efficient than the manually controlled WCE scanning, as the DRL-controlled WCE covered 98.04\% of the stomach area within 150.37 seconds.

\begin{table}[hbtp]
    \caption{Comparison between manual and DRL-based control}
    \centering
    \begin{tabular}{@{}ccc@{}}
    \toprule
    Time & Manual Control & DRL-based Control\\ \midrule
    60.02 seconds & 58.84\% & 72.87\% \\
    120.34 seconds & 80.68\% & 94.66\% \\
    150.37 seconds & 86.69\% & 98.04\% \\ \bottomrule
    \label{table2}
    \end{tabular}
\end{table}

\section{CONCLUSION AND FUTURE WORK}
\label{sec5}
In this work, We apply a virtual platform VR-Caps to simulate the process of stomach coverage scanning with a capsule endoscope model. Two monocular visual feedback-based DRL methods, i.e., the PPO and SAC, are utilized for training the permanent magnetic agent. We analyze the effect of the learning rate on training results; hence we get a set of appropriate hyperparameters for the training process. Finally, we can complete coverage of 72.87\%, 94.66\%, and 98.04\% of the stomach area in 60.02, 120.34, and 150.37 seconds, respectively. The coverage comparison between the manual and DRL control shows the efficiency of the DRL method.

Further work can be carried out in two aspects: (a) train and navigate the WCE in a more complex and challenging environment, such as the peristaltic irregularly shaped small intestine environment with food residues and air bubbles; (b) generalize the model to real scenarios.

\addtolength{\textheight}{-12cm}  





\section*{ACKNOWLEDGMENT}

We appreciate CUHK Research Postgraduate Student Grant for Overseas Academic Activities for supporting our conference travel. We thank K. İncetan \textit{et al.} \cite{c7} and A. Juliani \textit{et al.} \cite{c17} for open-sourcing their research work.


\bibliographystyle{IEEEtran}
\bibliography{reference}

\begin{thebibliography}{10}
\providecommand{\url}[1]{#1}
\csname url@samestyle\endcsname
\providecommand{\newblock}{\relax}
\providecommand{\bibinfo}[2]{#2}
\providecommand{\BIBentrySTDinterwordspacing}{\spaceskip=0pt\relax}
\providecommand{\BIBentryALTinterwordstretchfactor}{4}
\providecommand{\BIBentryALTinterwordspacing}{\spaceskip=\fontdimen2\font plus
\BIBentryALTinterwordstretchfactor\fontdimen3\font minus
  \fontdimen4\font\relax}
\providecommand{\BIBforeignlanguage}[2]{{%
\expandafter\ifx\csname l@#1\endcsname\relax
\typeout{** WARNING: IEEEtran.bst: No hyphenation pattern has been}%
\typeout{** loaded for the language `#1'. Using the pattern for}%
\typeout{** the default language instead.}%
\else
\language=\csname l@#1\endcsname
\fi
#2}}
\providecommand{\BIBdecl}{\relax}
\BIBdecl

\bibitem{c1}
J.~G. Albert, F.~Martiny, A.~Krummenerl, K.~Stock, J.~Lesske, C.~M. G{\"o}bel,
  E.~Lotterer, H.~H. Nietsch, C.~Behrmann, and W.~E. Fleig, ``Diagnosis of
  small bowel crohn’s disease: a prospective comparison of capsule endoscopy
  with magnetic resonance imaging and fluoroscopic enteroclysis,'' \emph{Gut},
  vol.~54, no.~12, pp. 1721--1727, 2005.

\bibitem{c6}
G.~Iddan, G.~Meron, A.~Glukhovsky, and P.~Swain, ``Wireless capsule
  endoscopy,'' \emph{Nature}, vol. 405, no. 6785, pp. 417--417, 2000.

\bibitem{c2}
G.~Ciuti, R.~Cali{\`o}, D.~Camboni, L.~Neri, F.~Bianchi, A.~Arezzo,
  A.~Koulaouzidis, S.~Schostek, D.~Stoyanov, C.~Oddo \emph{et~al.}, ``Frontiers
  of robotic endoscopic capsules: a review,'' \emph{Journal of micro-bio
  robotics}, vol.~11, pp. 1--18, 2016.

\bibitem{c15}
M.~Sitti, H.~Ceylan, W.~Hu, J.~Giltinan, M.~Turan, S.~Yim, and E.~Diller,
  ``Biomedical applications of untethered mobile milli/microrobots,''
  \emph{Proceedings of the IEEE}, vol. 103, no.~2, pp. 205--224, 2015.

\bibitem{c5}
C.~Hu, W.~Yang, D.~Chen, M.~Q.-H. Meng, and H.~Dai, ``An improved magnetic
  localization and orientation algorithm for wireless capsule endoscope,'' in
  \emph{2008 30th Annual International Conference of the IEEE Engineering in
  Medicine and Biology Society}.\hskip 1em plus 0.5em minus 0.4em\relax IEEE,
  2008, pp. 2055--2058.

\bibitem{c10}
A.~W. Mahoney and J.~J. Abbott, ``Five-degree-of-freedom manipulation of an
  untethered magnetic device in fluid using a single permanent magnet with
  application in stomach capsule endoscopy,'' \emph{The International Journal
  of Robotics Research}, vol.~35, no. 1-3, pp. 129--147, 2016.

\bibitem{c20}
L.~Bai, S.~Chen, M.~Gao, L.~Abdelrahman, M.~Al~Ghamdi, and M.~Abdel-Mottaleb,
  ``The influence of age and gender information on the diagnosis of diabetic
  retinopathy: based on neural networks,'' in \emph{2021 43rd annual
  international conference of the IEEE engineering in medicine \& Biology
  Society (EMBC)}.\hskip 1em plus 0.5em minus 0.4em\relax IEEE, 2021, pp.
  3514--3517.

\bibitem{c21}
A.~Smiti, ``When machine learning meets medical world: Current status and
  future challenges,'' \emph{Computer Science Review}, vol.~37, p. 100280,
  2020.

\bibitem{c23}
L.~Bai, L.~Wang, T.~Chen, Y.~Zhao, and H.~Ren, ``Transformer-based disease
  identification for small-scale imbalanced capsule endoscopy dataset,''
  \emph{Electronics}, vol.~11, no.~17, p. 2747, 2022.

\bibitem{c24}
H.~Che, H.~Jin, and H.~Chen, ``Learning robust representation for joint grading
  of ophthalmic diseases via adaptive curriculum and feature disentanglement,''
  in \emph{Medical Image Computing and Computer Assisted Intervention--MICCAI
  2022: 25th International Conference, Singapore, September 18--22, 2022,
  Proceedings, Part III}.\hskip 1em plus 0.5em minus 0.4em\relax Springer,
  2022, pp. 523--533.

\bibitem{c25}
Y.~Wu, S.~Zhao, S.~Qi, J.~Feng, H.~Pang, R.~Chang, L.~Bai, M.~Li, S.~Xia,
  W.~Qian \emph{et~al.}, ``Two-stage contextual transformer-based convolutional
  neural network for airway extraction from ct images,'' \emph{arXiv preprint
  arXiv:2212.07651}, 2022.

\bibitem{c12}
V.~Mnih, K.~Kavukcuoglu, D.~Silver, A.~Graves, I.~Antonoglou, D.~Wierstra, and
  M.~Riedmiller, ``Playing atari with deep reinforcement learning,''
  \emph{arXiv preprint arXiv:1312.5602}, 2013.

\bibitem{c13}
A.~Nair, B.~McGrew, M.~Andrychowicz, W.~Zaremba, and P.~Abbeel, ``Overcoming
  exploration in reinforcement learning with demonstrations,'' in \emph{2018
  IEEE international conference on robotics and automation (ICRA)}.\hskip 1em
  plus 0.5em minus 0.4em\relax IEEE, 2018, pp. 6292--6299.

\bibitem{c9}
B.~R. Kiran, I.~Sobh, V.~Talpaert, P.~Mannion, A.~A. Al~Sallab, S.~Yogamani,
  and P.~P{\'e}rez, ``Deep reinforcement learning for autonomous driving: A
  survey,'' \emph{IEEE Transactions on Intelligent Transportation Systems},
  vol.~23, no.~6, pp. 4909--4926, 2021.

\bibitem{c8}
L.~P. Kaelbling, M.~L. Littman, and A.~W. Moore, ``Reinforcement learning: A
  survey,'' \emph{Journal of artificial intelligence research}, vol.~4, pp.
  237--285, 1996.

\bibitem{c4}
P.~Henderson, R.~Islam, P.~Bachman, J.~Pineau, D.~Precup, and D.~Meger, ``Deep
  reinforcement learning that matters,'' in \emph{Proceedings of the AAAI
  conference on artificial intelligence}, vol.~32, no.~1, 2018.

\bibitem{c14}
J.~Schulman, F.~Wolski, P.~Dhariwal, A.~Radford, and O.~Klimov, ``Proximal
  policy optimization algorithms,'' \emph{arXiv preprint arXiv:1707.06347},
  2017.

\bibitem{c3}
T.~Haarnoja, A.~Zhou, P.~Abbeel, and S.~Levine, ``Soft actor-critic: Off-policy
  maximum entropy deep reinforcement learning with a stochastic actor,'' in
  \emph{International conference on machine learning}.\hskip 1em plus 0.5em
  minus 0.4em\relax PMLR, 2018, pp. 1861--1870.

\bibitem{c19}
G.~Trovato, M.~Shikanai, G.~Ukawa, J.~Kinoshita, N.~Murai, J.~Lee, H.~Ishii,
  A.~Takanishi, K.~Tanoue, S.~Ieiri \emph{et~al.}, ``Development of a colon
  endoscope robot that adjusts its locomotion through the use of reinforcement
  learning,'' \emph{International journal of computer assisted radiology and
  surgery}, vol.~5, pp. 317--325, 2010.

\bibitem{c18}
L.~Wu, J.~Zhang, W.~Zhou, P.~An, L.~Shen, J.~Liu, X.~Jiang, X.~Huang, G.~Mu,
  X.~Wan \emph{et~al.}, ``Randomised controlled trial of wisense, a real-time
  quality improving system for monitoring blind spots during
  esophagogastroduodenoscopy,'' \emph{Gut}, vol.~68, no.~12, pp. 2161--2169,
  2019.

\bibitem{c16}
M.~Turan, Y.~Almalioglu, H.~B. Gilbert, F.~Mahmood, N.~J. Durr, H.~Araujo,
  A.~E. Sar{\i}, A.~Ajay, and M.~Sitti, ``Learning to navigate endoscopic
  capsule robots,'' \emph{IEEE Robotics and Automation Letters}, vol.~4, no.~3,
  pp. 3075--3082, 2019.

\bibitem{c11}
A.~Marino, B.~Scaglioni, and P.~Valdastri, ``Reinforcement learning based
  control for a magnetic flexible endoscope,'' in \emph{Proceedings of the
  Hamlyn Symposium on Medical Robotics}, 2021.

\bibitem{c7}
K.~{\.I}ncetan, I.~O. Celik, A.~Obeid, G.~I. Gokceler, K.~B. Ozyoruk,
  Y.~Almalioglu, R.~J. Chen, F.~Mahmood, H.~Gilbert, N.~J. Durr \emph{et~al.},
  ``Vr-caps: a virtual environment for capsule endoscopy,'' \emph{Medical image
  analysis}, vol.~70, p. 101990, 2021.

\bibitem{c17}
A.~Juliani, V.-P. Berges, E.~Teng, A.~Cohen, J.~Harper, C.~Elion, C.~Goy,
  Y.~Gao, H.~Henry, M.~Mattar \emph{et~al.}, ``Unity: A general platform for
  intelligent agents,'' \emph{arXiv preprint arXiv:1809.02627}, 2018.

\end{thebibliography}

\end{document}